\documentclass[10pt,twocolumn,letterpaper]{article}

\usepackage[pagenumbers]{cvpr} %

\usepackage{graphicx}
\usepackage{amsmath}
\usepackage{amssymb}
\usepackage{booktabs}
\usepackage{circledsteps}

\usepackage[pagebackref,breaklinks,colorlinks]{hyperref}

\usepackage[capitalize]{cleveref}
\crefname{section}{Sec.}{Secs.}
\Crefname{section}{Section}{Sections}
\Crefname{table}{Table}{Tables}
\crefname{table}{Tab.}{Tabs.}

\def\cvprPaperID{1793} %
\def\confName{CVPR}
\def\confYear{2022}

\usepackage{overpic}
\usepackage{multirow}
\usepackage{microtype}
\usepackage{enumitem}
\usepackage{overpic}
\usepackage[below]{placeins}
\usepackage{dsfont}
\usepackage{wrapfig}
\usepackage{booktabs}
\usepackage{balance}
\usepackage[rflt]{floatflt}
\usepackage{stfloats} %

\usepackage{color}
\definecolor{turquoise}{cmyk}{0.65,0,0.1,0.3}
\definecolor{purple}{rgb}{0.65,0,0.65}
\definecolor{dark_green}{rgb}{0, 0.5, 0}
\definecolor{orange}{rgb}{0.8, 0.6, 0.2}
\definecolor{red}{rgb}{0.8, 0.2, 0.2}
\definecolor{darkred}{rgb}{0.6, 0.1, 0.05}
\definecolor{blueish}{rgb}{0.0, 0.3, .6}
\definecolor{light_gray}{rgb}{0.7, 0.7, .7}
\definecolor{pink}{rgb}{1, 0, 1}
\definecolor{greyblue}{rgb}{0.25, 0.25, 1}
\definecolor{gold}{rgb}{0.7, 0.5, 0}

\newcommand{\TODO}[1]{}

\newcommand{\At}[1]{}
\newcommand{\AT}[1]{}

\newcommand{\dr}[1]{{#1}}
\newcommand{\Dr}[1]{}
\newcommand{\DR}[1]{}

\newcommand{\MM}[1]{}

\newcommand{\DL}[1]{}

\newcommand{\KY}[1]{}

\newcommand{\SupplementaryMaterial}{{supplementary material}\xspace}

\usepackage{blindtext}

\newcommand{\Figure}[1]{Figure~\ref{fig:#1}}

\newcommand{\Table}[1]{Table~\ref{tab:#1}}
\newcommand{\eq}[1]{\eqref{eqn:#1}}

\newcommand{\Equation}[1]{Equation~\eqref{eqn:#1}}

\newcommand{\Section}[1]{Section~\ref{sec:#1}}

\DeclareMathOperator*{\argmin}{arg\,min}

\newcommand{\loss}[1]{\mathcal{L}_\text{#1}}
\newcommand{\expect}{\mathbb{E}}
\newcommand{\real}{\mathbb{R}}

\newcommand{\probP}{\text{I\kern-0.15em P}}

\usepackage{silence}
\renewcommand{\paragraph}[1]{\vspace{.2em}\noindent\textbf{#1}.}

\renewcommand{\Circled}[1]{\raisebox{.5pt}{\footnotesize \textcircled{\raisebox{-.6pt}{#1}}}}

\usepackage{lipsum}

\newcommand{\methodname}{LOLNeRF\xspace}
\newcommand{\pigan}{$\pi$-GAN~\cite{Chan21cvpr_piGAN}\xspace}
\newcommand{\symmetry}{UnSup3D~\cite{Wu2020ulp}\xspace}

\newcommand{\position}{\mathbf{x}}
\newcommand{\depth}{d}
\newcommand{\viewdir}{\mathbf{d}}
\newcommand{\density}{\sigma}
\newcommand{\radiance}{c}
\newcommand{\posenc}{\mathbf{\gamma}}
\newcommand{\weight}{w}
\newcommand{\latent}{\mathbf{z}}
\newcommand{\latenttable}{\mathbf{Z}}
\newcommand{\latentdim}{D}
\newcommand{\tablesize}{K}
\newcommand{\latentdistribution}{\mathcal{Z}}
\newcommand{\latentmean}{\mu_\mathbf{Z}}
\newcommand{\latentcovariance}{\chi_\mathbf{Z}}
\newcommand{\pixelcolor}{C}
\newcommand{\backgroundcolor}{C_\textrm{bg}}
\newcommand{\nerfcolor}{C_\textrm{NeRF}}
\newcommand{\pixelalpha}{\alpha}
\newcommand{\pixeldepth}{D}
\newcommand{\image}{I}
\newcommand{\pixel}{\mathbf{p}}
\newcommand{\segmenter}{S}
\newcommand{\lossweight}{\lambda}

\begin{document}

\title{\includegraphics[width=12pt]{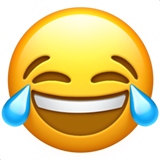}LOLNeRF: Learn from One Look}

\author{Daniel Rebain \textsuperscript{1,3}
\and
Mark Matthews\textsuperscript{3}
\and
Kwang Moo Yi\textsuperscript{1}
\and 
Dmitry Lagun\textsuperscript{3}
\and
Andrea Tagliasacchi\textsuperscript{2,3}
\vspace{0.5em}
\\
\vspace{-2em}
\textsuperscript{1}University of British Columbia $\quad$
\textsuperscript{2}Simon Fraser University $\quad$
\textsuperscript{3}Google Research}

\twocolumn[{
\renewcommand\twocolumn[1][]{#1}
\maketitle
\begin{center}
\centering
\captionsetup{type=figure}
\includegraphics[width=\linewidth, trim= 0 0 0 5, clip]{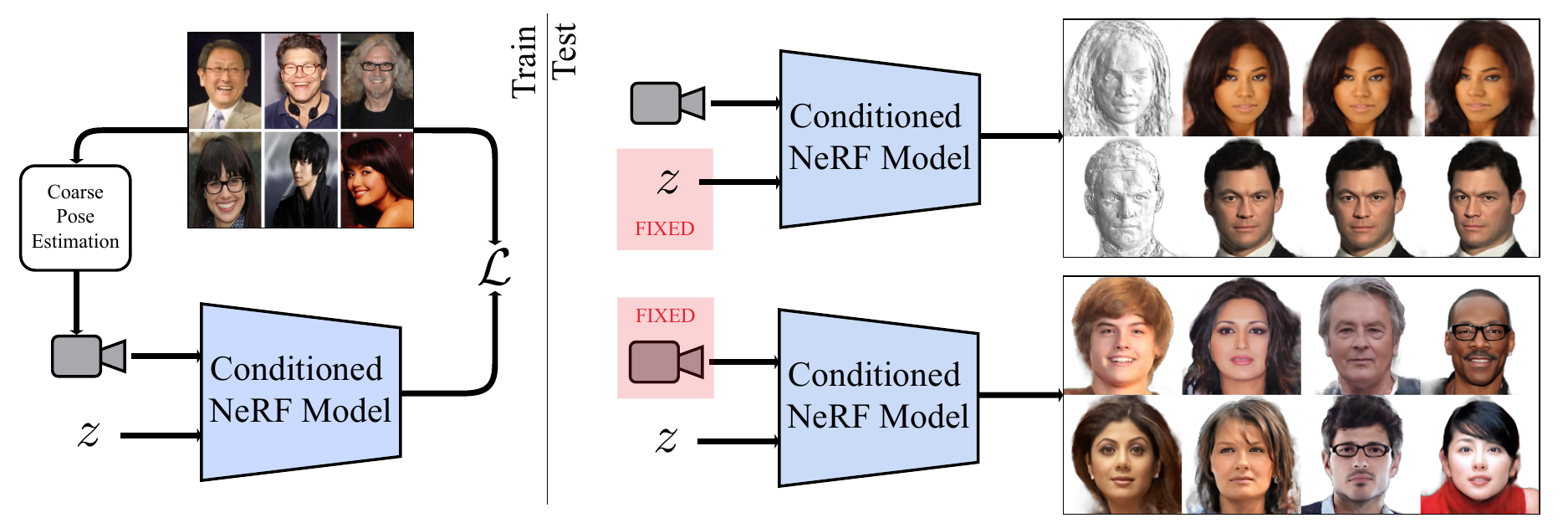}
\vspace{-2em}
\captionof{figure}{
\textbf{Teaser} --
Our method learns a space of shape and appearance by reconstructing a large collection of \textit{single-view} images using a single neural network conditioned on a \textit{shared} latent space (left).
This allows for a volumetric 3D model to be lifted from the image, and rendered from novel viewpoints (right).
\url{https://lolnerf.github.io}
}
\label{fig:teaser}

\end{center}
}]

\setlength\belowcaptionskip{-0.5ex}
\begin{abstract}
   We present a method for learning a generative 3D model based on neural radiance fields, trained solely from data with only single views of each object.
While generating realistic images is no longer a difficult task, producing the corresponding 3D structure such that they can be rendered from different views is non-trivial.
We show that, unlike existing methods, one does not need multi-view data to achieve this goal.
Specifically, we show that by reconstructing many images aligned to an approximate canonical pose with a single network conditioned on a shared latent space, you can learn a space of radiance fields that models shape and appearance for a class of objects.
We demonstrate this by training models to reconstruct object categories using datasets that contain only one view of each subject without depth or geometry information.
Our experiments show that we achieve state-of-the-art results in novel view synthesis and high-quality results for monocular depth prediction.
\url{https://lolnerf.github.io}

\end{abstract}
\vspace{-2em}
\section{Introduction}
\label{sec:intro}

A long-standing challenge in computer vision is the extraction of 3D geometric information from images of the real world~\cite{Malik2016threer}.
Understanding 3D geometry is critical to understanding the physical and semantic structure of objects and scenes, but achieving it remains a very challenging problem.
Work in this area has mainly focused either on deriving geometric understanding from more than one view~\cite{Andrew2001mvg, Sitzmann19neurips_srn, Jang21iccv_CodeNeRF}, or by using known geometry to supervise the learning of geometry from single views~\cite{Mescheder19cvpr_occupancy_net, Deng20cvpr_cvxnet, Chen20cvpr_bsp, Genova19iccv_sif}.
Here, we take a more ambitious approach and aim to derive equivalent 3D understanding in a generative model from only \textit{single} views of objects, and without relying on explicit geometric information like depth or point clouds.
Deriving such 3D understanding is, however, non trivial.
While Neural Radiance Field (NeRF)-based methods~\cite{Mildenhall20eccv_nerf, Dellaert2021nerfexplosion} have shown great promise in geometry-based rendering, they focus on learning a single scene from multiple views.

Existing NeRF works \cite{Mildenhall20eccv_nerf, Park21iccv_nerfies, Gafni21cvpr_DNRF} all require supervision from more than one viewpoint, as without it, NeRF methods are prone to collapse to a flat representation of the scene, because they have no incentive to create a volumetric representation; see~\Figure{mainpoint}~(left).
This serves as a major bottleneck, as multiple-view data is hard to acquire.
Thus, architectures have been devised to workaround this that combine NeRF and Generative Adversarial Networks (GANs)~\cite{Schwarz20neurips_graf, Chan21cvpr_piGAN, Niemeyer2021campari}, where the multi-view consistency is enforced through a discriminator to avoid the need for multi-view training data.

\begin{floatingfigure}{.5\linewidth}
\centering
\begin{overpic}
[width=.45\linewidth]
{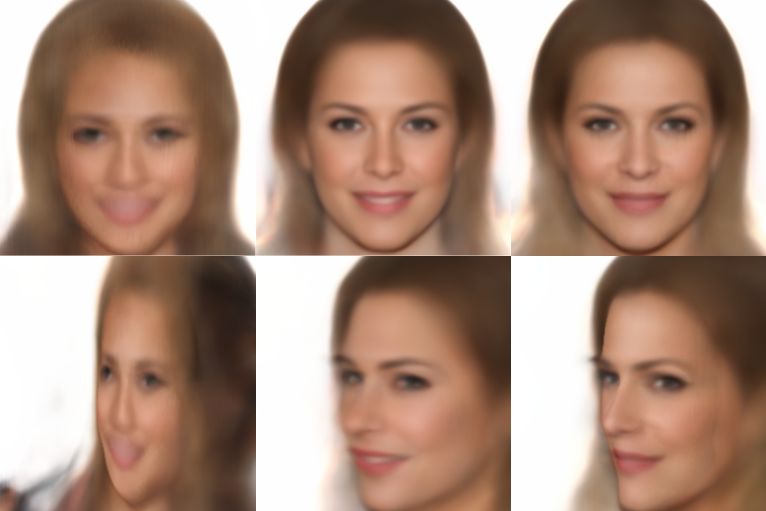}
\put(-7,40){\footnotesize\rotatebox{90}{train}}
\put(-7,11){\footnotesize\rotatebox{90}{test}}
\put(7, -7){\footnotesize{(left)}}
\put(37,-7){\footnotesize{(middle)}}
\put(74,-7){\footnotesize{(right)}}
\end{overpic}
\captionof{figure}{}
\label{fig:mainpoint}
\vspace{-1em}
\end{floatingfigure}
In this work we show that -- surprisingly -- having only single views of a class of objects is enough to train NeRF models without adversarial supervision, as long as a \textit{shared} generative model is trained, and \textit{approximate} camera poses are provided.
In a nutshell, the multi-view constraint of existing works no longer necessarily needs to be enforced, and cameras do not have to be accurate to achieve compelling results; see~\Figure{mainpoint}~(right).
Specifically, we roughly align all images in the dataset to a canonical pose using predicted 2D landmarks, which is then used to determine from which view the radiance field should be rendered to reproduce the original image.
For the generative model we employ an auto-decoder framework~\cite{Park19cvpr_deepsdf}.
To improve generalization, we further train two models, one for the foreground -- the common object class of the dataset -- and one for the background, since the background is often inconsistent throughout the data, hence~unlikely to be subject to the 3D-consistency bias.
We also encourage our model to model shapes as solid surfaces (i.e.~sharp outside-to-inside transitions), which further improves the quality of predicted shapes; see the improvements from \Figure{mainpoint}~(middle) to \Figure{mainpoint}~(right).

A noteworthy aspect of our method is that we \textit{do not} require rendering of entire images, or even patches, while training. 
In the auto-decoder framework, we train our models to reconstruct images from datasets, and at the same time find the optimal latent representations for each image -- an objective that can be enforced on \textit{individual} pixels.
Hence, our method can be trained with arbitrary image sizes without any increase in memory requirement during training.
In contrast, existing methods that utilize GANs~\cite{Schwarz20neurips_graf, Chan21cvpr_piGAN, Niemeyer2021campari} supervise on inter-pixel relationships through their discriminators, greatly limiting or outright preventing them from being able to scale with respect to training image resolution.
\begin{figure}[t]
\centering
\begin{overpic}
[width=\linewidth]
{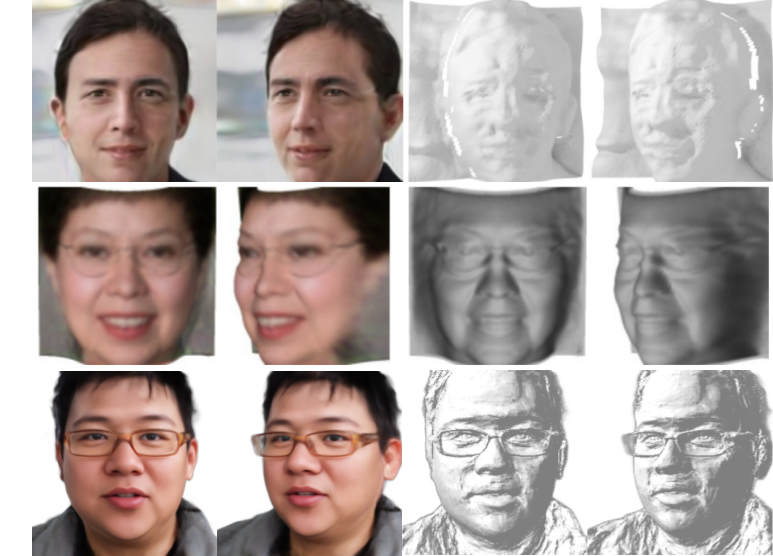}
\put(0,3){\footnotesize\rotatebox{90}{Our Method}}
\put(0,27){\footnotesize\rotatebox{90}{Wu et al.~\cite{Wu2020ulp}}}
\put(0,51){\footnotesize\rotatebox{90}{\pigan}}
\end{overpic}
\vspace{-2em}
\caption{
\textbf{Geometry recovery} --
Our model is able to capture reasonable coarse geometry from a single view as well as topologically complex detail like glasses that can't be encoded in a depth map.
}
\vspace{-1em}
\label{fig:qual_geometry}
\end{figure}
In summary, we:
\vspace{-.5em}
\begin{itemize}[leftmargin=*]
    \setlength\itemsep{-.3em}
    \item propose a method for learning 3D reconstruction of object categories from \textit{single-view} images \dr{which decouples training complexity from image resolution};
    \item show that single views are enough to learn high-quality prediction of geometry (e.g.~depth) without any geometric supervision (\Figure{qual_geometry});    
    \item show that our method exceeds adversarial methods in representing appearance of objects from the learned category by reconstructing held-out images and novel views.
\end{itemize}

\section{Related Work}

There are three main research topics related to ours:~\Circled{1}~classic 3DMMs, \Circled{2} {neural implicit shape representations}, and  \Circled{3} {shape estimation from single view}.
We also review~NeRF and auto-decoders, on which we base our research.

\paragraph{3D Morphable Models~(3DMM)}
Classical approaches to building shape spaces have focused on parameterized 3D mesh representations, with much work devoted to human faces~\cite{Blanz1999mms, Paysan2009tdf, Gerig2018mfm}, as surveyed by Egger et al.\cite{Egger2020tdm}.
These models are typically built from geometric information sources like 3D scans or depth sensors, which are highly accurate, but require significant investment (e.g. the proprietary Disney Medusa capture rig\cite{Riviere2020ssh,Gotardo2018pdf} or the commercially available solutions provided by 3DMD\cite{Threedmd2021products}).
The use of mesh representations also limits these models for applications like novel view synthesis where phenomena like hair are not well reproduced.
By contrast, our approach is relatively unconstrained, as it derives flexible volumetric representations directly from images taken in uncontrolled environments.

\paragraph{Neural Implicit Representations}
Representing scenes as 3D implicit fields has proven successful for a number of tasks.
These models can take a number of forms, including representations of distance~\cite{Park19cvpr_deepsdf}, occupancy~\cite{Mescheder19cvpr_occupancy_net,Chen2019imnet}, learned scene features~\cite{Sitzmann19neurips_srn}, and light fields~\cite{Sitzmann2021lfn}.
One representation in particular has been highly successful in simultaneously modeling shape and appearance: radiance fields.
Neural Radiance Field models, known as NeRFs~\cite{Mildenhall20eccv_nerf}, use fields for density and radiance, and are particularly effective in learning 3D scene structure from images alone.
A substantial number of extensions to NeRF have been proposed~\cite{Dellaert2021nerfexplosion}, some notable examples being: per-view appearance codes \cite{MartinBrualla21cvpr_nerfw}, multi-resolution training \cite{Barron21iccv_Mip_NeRF}, camera co-optimization \cite{Wang21arxiv_NeRFminusminus, Lin21iccv_BARF, Niemeyer2021campari}, hard surface priors\cite{Oechsle21iccv_UNISURF}, deformable scenes \cite{Park21iccv_nerfies}, variable topology \cite{Park2021hypernerf} and foreground-background decompositions\cite{Zhang20arxiv_NeRFplusplus}.
The single-scene formulation of NeRF has also been extended to general object classes with~\cite{Jang21iccv_CodeNeRF, Xie2021fignerf}, and hybridized with GAN methods as in GNeRF\cite{Meng21iccv_GNeRF}, GIRAFFE~\cite{Niemeyer21cvpr_GIRAFFE}, and StyleNeRF\cite{Anonymous2021stylenerf}.

\paragraph{Shape from Single View}
A long-standing objective in computer vision has been to understand the 3D structure of scenes and objects from a single image.
Many works have approached this problem by encoding a relationship between appearance and structure using prior knowledge of that structure as supervision~\cite{Genova19iccv_sif, Deng20cvpr_cvxnet, Mescheder19cvpr_occupancy_net, Henzler21cvpr_object_categories}.
Until recently however, the problem of deriving such a model from only single-view observations has remained very difficult.
Wu et al. \cite{Wu2020ulp} demonstrate how shape can be inferred for categories of objects which are approximately symmetric.

The most relevant to our work however, has been the development of GAN-based methods which learn a space of shapes that when rendered produce a distribution of images is indistinguishable from a training distribution~\cite{Schwarz20neurips_graf, Chan21cvpr_piGAN, Niemeyer2021campari, Pan2021shadegan}.
This approach is effective in producing models with plausible structure, as they enforce an implicit multi-view constraint by requiring that renders from any viewpoint appear realistic.
Unfortunately, this requires the use of discriminator networks which are very inefficient when combined with 3D representations that use volumentric representations.
To avoid this limitation, we reconstruct images directly with a more efficient and scalable stochastic sampling process.

\paragraph{Neural Radiance Fields~(NeRF)}
Neural Radiance Fields~\cite{Mildenhall20eccv_nerf} use classical volume rendering~\cite{Kajiya1984rtv} to compute radiance values for each pixel~$\pixel$ from samples taken at points $\position$ along the associated ray.
These samples are computed using a learned \emph{radiance field} which maps $\position$, as well as the ray direction $\viewdir$, to radiance values $\radiance$ and density values~$\density$.
The volume rendering equation takes the form a weighted sum of the radiance values at each sample point~$\position_i$:
\vspace{-0.8em}
\begin{equation}
\nerfcolor(\pixel) = \sum_{i=1}^N \weight_i \cdot \radiance(\position_i, \viewdir)
\vspace{-0.8em}
\end{equation}
with the weights $\weight_i$ being derived from an accumulation of the transmittance along the view ray~$\position_i$:
\vspace{-0.8em}
\begin{equation}
\label{eqn:sample_weight}
\weight_i = \left(1 - \mathrm{exp}(-\density(\position_i) \delta_i) \right)
\,\cdot\,
\mathrm{exp}\left(-\sum_{j=1}^{i-1} \density(\position_j) \delta_j\right)
\vspace{-0.8em}
\end{equation}
where $\delta_i$ is the sample spacing at the $i$-th point.
Note here that we denote the product of the accumulated transmittance and sample opacity as $\weight$, as this value determines the contribution of a single sample to the final pixel value.
These weights can also be used to compute other values such as surface depth (by replacing the per sample radiance values with sample depth $\depth(\position_i)$, or the overall pixel opacity:
\vspace{-0.8em}
\begin{equation}
\pixeldepth(\pixel) = \sum_{i=1}^N \weight_i \cdot \depth(\position_i) \quad \quad \pixelalpha(\pixel) = \sum_{i=1}^N \weight_i \cdot 1
\vspace{-0.8em}
\end{equation}

\paragraph{Auto-decoders}
Auto-decoders~\cite{Park19cvpr_deepsdf, Tan1995rdd, Sitzmann20neurips_metaSDF}, also known as Generative Latent Optimization (GLO)~\cite{Bojanowski2019ols, MartinBrualla21cvpr_nerfw, Park19cvpr_deepsdf, Singh2019gloss, Azuri2020gli}, are a family of generative models that learn without the use of either an encoder or discriminator.
The method works similarly to an auto-\emph{encoder}, in that a decoder network maps a latent code to a final output.
However the method differs in how these latent codes are found: auto-decoders learn the codes directly by allocating 
a table of codes with a row 
for each distinct element in the training dataset.
These codes are \textit{co-optimized} with the rest of the model parameters as learnable variables.
\begin{figure}[t]
\centering
\begin{overpic}
[width=\linewidth, trim = 0 185 0 0, clip]
{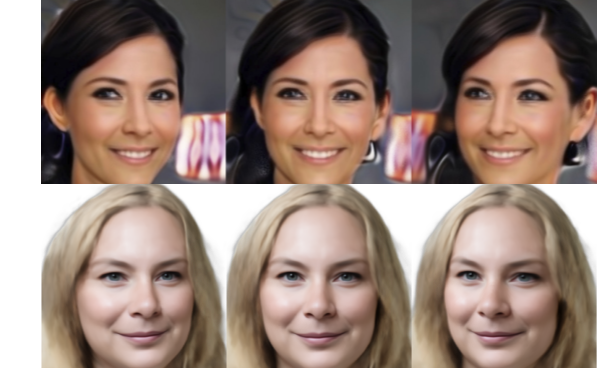}
\put(2,5){\footnotesize\rotatebox{90}{\pigan}}
\end{overpic}

\vspace{0.5em}
\begin{overpic}
[width=\linewidth, trim = 0 0 0 185, clip]
{fig/novel.png}
\put(2,5){\footnotesize\rotatebox{90}{Our Method}}
\end{overpic}
\vspace{-1em}
\caption{\textbf{Novel view synthesis} -- 
For each method we show an example appropriate to the training method: for \pigan~a latent code sampled from the training distribution, and for ours a learned latent code which reconstructs a training image.
Our method recovers more sharp detail due to training on higher resolution images.
For a comparison of novel views of the same image reconstructed by both methods,~see~\Figure{qual_fitting}.
}
\vspace{-1em}
\label{fig:qual_novel_view}
\end{figure}
\begin{figure*}[t]
\centering
\includegraphics[width=1.0\linewidth, trim=0 0 0 5, clip]{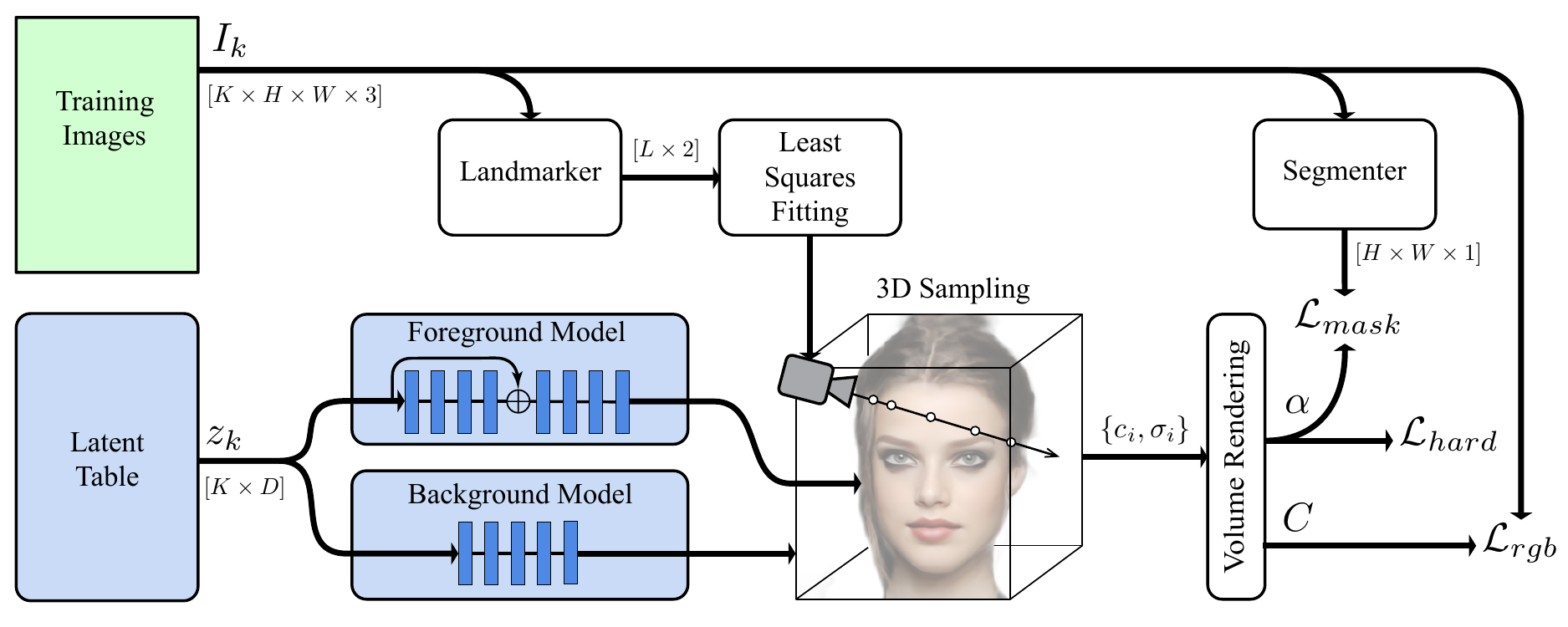}
\vspace{-2em}
\definecolor{input}{RGB}{211, 254, 211}
\definecolor{learned}{RGB}{203, 219, 246}
\caption{\textbf{Architecture} -- Our method learns a per-image table of latent codes alongside foreground and background NeRFs. Volumetric rendering output is subject to a per-ray RGB loss against each training pixel, and alpha value against an image segmenter.  Camera alignments are derived from a least-squares fit of 2D landmarker outputs to class-specific canonical 3D keypoints. 
\\\hspace{\textwidth}
\textbf{Legend:} \colorbox{input}{input data} \colorbox{learned}{learned} \framebox{fixed function}
}
\vspace{-1em}
\label{fig:architecture}
\end{figure*}
\section{Method}
\label{sec:method}

We visualize our architecture in \Figure{architecture}.
We train our network parameters and latent codes $\latenttable$ by minimizing the weighted sum of three losses:
\vspace{-0.8em}
\begin{equation}
\loss{total} = \loss{rgb} + \underbrace{\lossweight_\textrm{mask} \loss{mask}}_\text{\Section{mask}} + \underbrace{\lossweight_\textrm{hard} \loss{hard}}_\text{\Section{hard}}
\vspace{-0.8em}
\end{equation}
where the first term is the standard L2 photometric reconstruction loss over pixels $\pixel$ from the training images $\image_k$:
\vspace{-0.8em}
\begin{equation}
\loss{rgb} = \expect_{k \in \{1..K\}, \pixel \in \image_k} \left[ (\pixelcolor(\pixel | \latent_k) - \pixelcolor_\mathrm{GT}^k(\pixel))^2 \right]
\end{equation}
We extend the ``single-scene'' (i.e. overfitting/memorization) formulation of NeRF to support learning a \textit{latent space of shapes} by incorporating an auto-decoder architecture.~$\quad$~In this modified architecture, the main NeRF backbone network is conditioned on a per-object latent code $\latent \in \real^\latentdim$, as well as the $L$-dimensional positional encoding $\posenc^L(x)$ as in \cite{Mildenhall20eccv_nerf}.
Mathematically, the density and radiance functions are then of the form $\density(\position | \latent)$ and $\radiance(\position | \latent)$; note we consider a formulation where radiance is not a function of view direction~$\viewdir$.~$\quad$~These latent codes are rows from the latent table $\latenttable {\in} \real^{\tablesize \times \latentdim}$, which we initialize to $\textbf{0}^{\tablesize \times \latentdim}$, where $K$ is the number of images.
This architecture makes it possible to accurately reconstruct training examples without requiring significant extra computation and memory for an encoder model, and avoids requiring a convolutional network to extract 3D information from the training images~\cite{Trevithick21iccv_GRF,Yu21cvpr_pixelNeRF}.
Training this model follows the same procedure as single-scene NeRF, but draws random rays from all $\tablesize$ images in the dataset, and associates each ray with the latent code that corresponds to the object in the image it was sampled from.

\subsection{Foreground-Background Decomposition}
\label{sec:mask}
Similar to \cite{Zhang20arxiv_NeRFplusplus, Niemeyer2021campari, Gao21iccv_DynamicVS}, we use a separate model to handle the generation of background details.
We use a lower-capacity model~$\backgroundcolor(\viewdir | \latent)$ for the background that predicts radiance on a per-ray basis.
We then render by combining the background and foreground colors using a transparency value derived from the NeRF density function:
\begin{align}
\pixelcolor(\pixel | \latent) &=
\pixelalpha(\pixel | \latent) \cdot \nerfcolor(\pixel | \latent) \\
&+(1 - \pixelalpha(\pixel | \latent))
\cdot \backgroundcolor(\viewdir_\pixel | \latent) \nonumber
\label{eqn:background}
\end{align}
In practice supervising the foreground/background separation is not always necessary; 
see the SRN Cars~\cite{Sitzmann19neurips_srn} results in \Figure{qual_renders} which learn a foreground decomposition naturally from solid background color and 360$^{\circ}$ camera distribution.
When a pre-trained module is available for predicting the \textit{foreground segmentation} of the training images, we also apply an additional loss to encourage the transparency of the NeRF volume to be consistent with this prediction:
\begin{equation}
\label{eqn:mask}
\loss{mask} = \expect_{k \in \{1..K\}, \pixel \in \image_k} \left[ ( \pixelalpha(\pixel | \latent_k) - \segmenter_\image(\pixel) )^2 \right]
\end{equation}
where $\segmenter_\image(\cdot)$ is the pre-trained image segmenter applied to image $\image_k$ and sampled at pixel $\pixel$.
When training on face datasets, we employ the \textit{MediaPipe} Selfie Segmentation~\cite{Google2021ss} for the pre-trained module in~\eq{mask} and $\lossweight_\textrm{mask} {=} 1.0$.

\subsection{Hard Surfaces}
\label{sec:hard}
NeRF does not explicitly enforce that the learned volumetric function strictly model a hard surface.
With enough input images, and sufficiently textured surfaces, multi-view consistency 
will favor the creation of hard transitions from empty to solid space.
Unfortunately, this property does not 
hold in the single view case.
Because the field function that corresponds to each latent code is only supervised from \textit{one} viewpoint, this often results in blurring of the surface \textit{along} the view direction; see \Figure{mainpoint}.
To counter this, we impose a prior on the probability of the weights $\weight$ to be distributed as a mixture of Laplacian distributions, one with mode around weight zero, and one with mode around weight one:
\begin{equation}
\probP(\weight) \propto e^{-|\weight|} + e^{-|1-\weight|}
\end{equation}
Note the distribution is peaky, and will encourage a sparse solution where any values of $\weight$ in the open interval $(0,1)$ to be discouraged.
We convert this prior into a loss via: 
\begin{equation}
    \loss{hard} = -\log(\probP(\weight))
    \label{eqn:hard2}
\end{equation}
The magnitude of $\density(\position)$ which satisfies this constraint depends on the sampling density. \Equation{hard2} encourages the density to produce a step function that saturates sampling weight over at least one sampling interval, which, by construction, is appropriate for the scale of scene being modelled.
We employ $\lossweight_\textrm{hard} = 0.1$ in our experiments.

\begin{figure}
\centering
\includegraphics[width=1.0\linewidth]{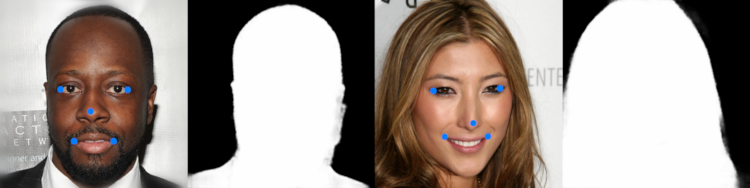}
\vspace{-1.5em}
\definecolor{landmarker}{RGB}{69, 126, 247}
\caption{\textbf{Landmarker and Segmenter} -- 
(Samples outputs from the landmarker and the segmenter networks for two input identities. Blue circles (\textcolor{landmarker}{$\bullet$}) denote the identified landmarks.
}
\label{fig:landmarker_segmenter}
\end{figure}

\subsection{Camera Parameters}
Volume rendering requires camera parameters that associate each pixel with a ray used to compute sample locations.
In classic NeRF, cameras are estimated by structure-from-motion on the input image dataset.
For our single-view use case, this is not possible due to depth ambiguity.
To make our method compatible with single-view images, we
we employ the \textit{MediaPipe} Face Mesh\cite{Google2021fm} pre-trained network module to extracts 2D landmarks that appear in consistent locations for the object class being considered. \Figure{landmarker_segmenter} shows example network output of the five landmarks used for human faces.

These landmark locations are then aligned with projections of canonical 3D landmark positions with a ``shape matching'' least-squares optimization to acquire a rough estimate of camera parameters; see the \SupplementaryMaterial for more detail on this process.

\begin{figure*}[t]
\hspace{1em}
\begin{overpic}
[width=\linewidth]
{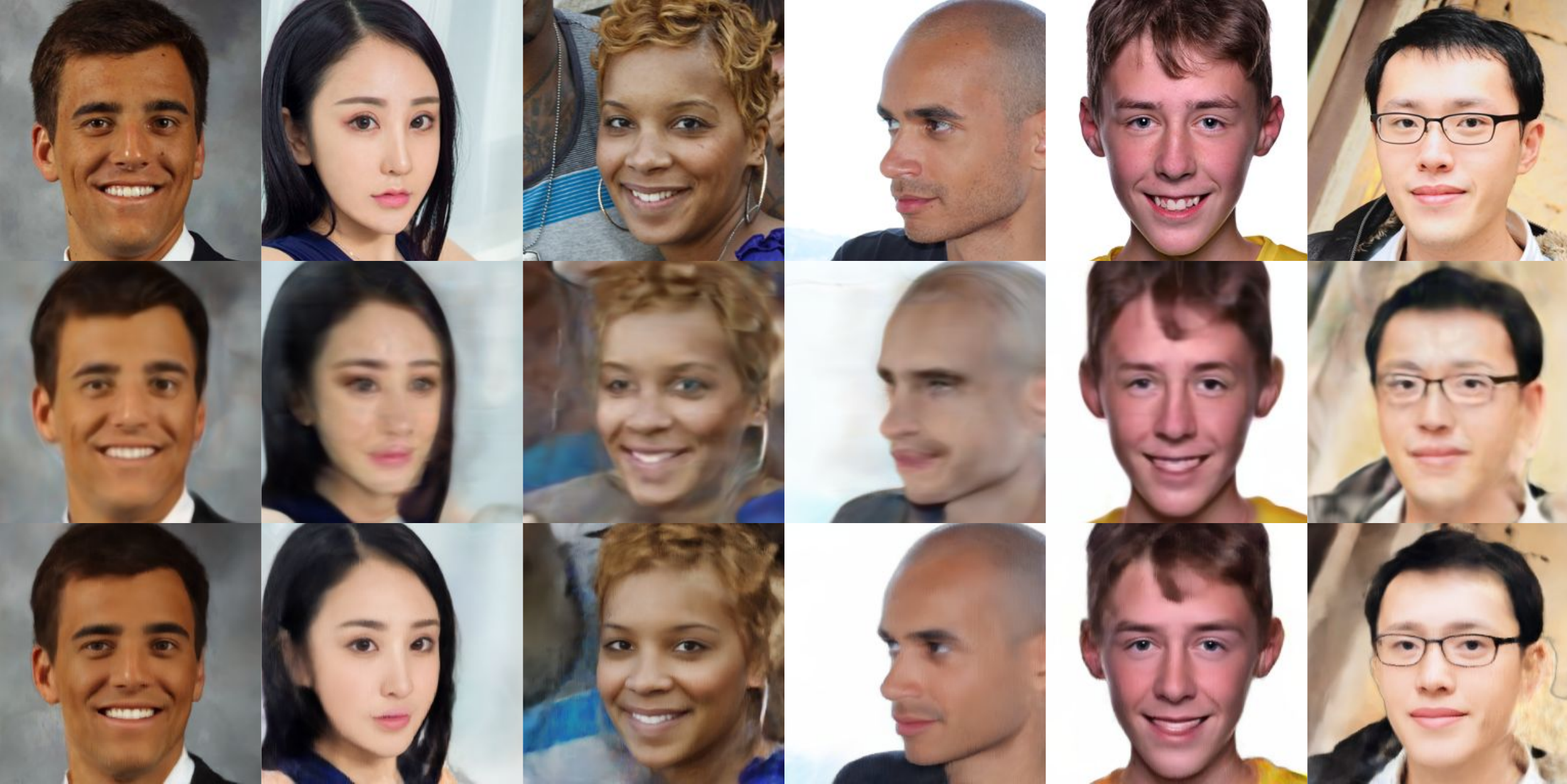}
\put(-2,38){\scriptsize\rotatebox{90}{FFHQ Target}}
\put(-2,22){\scriptsize\rotatebox{90}{$\pi$-GAN Fit}}
\put(-2,4){\scriptsize\rotatebox{90}{LOLNeRF Fit}}
\end{overpic}
\caption{
\textbf{Conditional generation} -- we show examples of both our method (trained on CelebA-HQ@$256^2$) and $\pi$-GAN (trained on CelebA@$128^2$) fitting to images not seen during training.
Our method produces much sharper reconstructions, especially for non-frontal views and appearances that are not well represented in the training set.
}
\label{fig:supp_fitting}
\end{figure*}

\begin{figure}[t]
\centering
\hspace{0.1\linewidth}
\begin{overpic}[width=0.9\linewidth, trim=50 0 0 0, clip]
{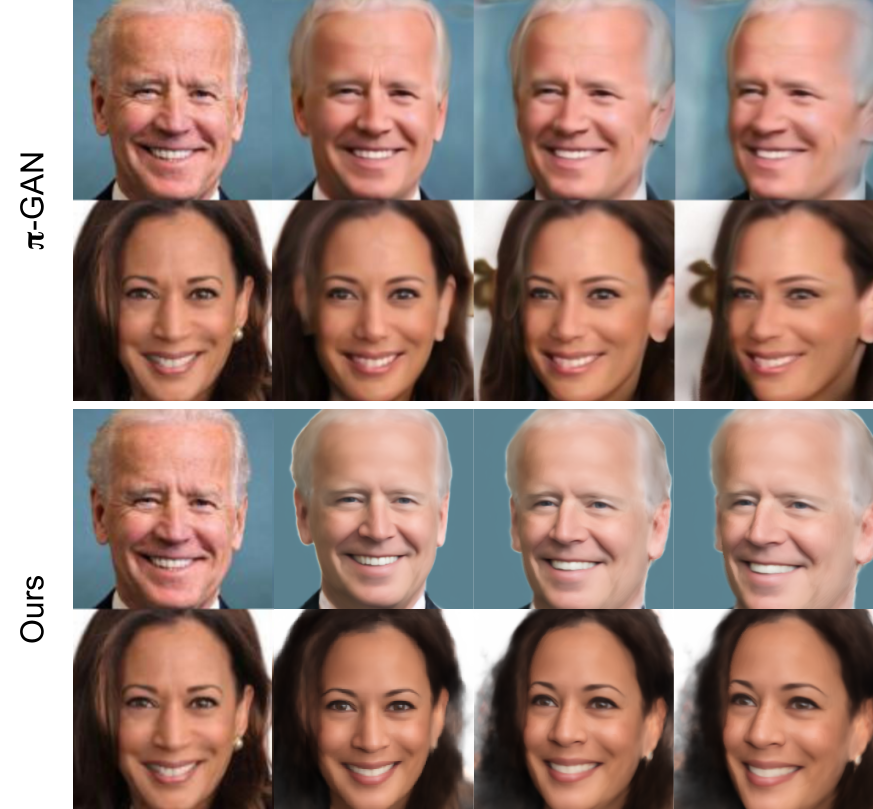}
\put(-2,15){\footnotesize\rotatebox{90}{Our Method}}
\put(-2,62){\footnotesize\rotatebox{90}{\pigan}}
\end{overpic}
\vspace{-0.5em}
\caption{
\textbf{Novel view comparison} --
Comparison of novel views on held-out image fitting with our method and \pigan.
Our method captures more fine detail and produces sharper results at large angles.
}
\vspace{-0.5em}
\label{fig:qual_fitting}
\end{figure}

\subsection{Conditional Generation}
\label{sec:fitting}
Given a pre-trained model, we can find a latent code $z$ which reconstructs an image which was not present in the training set.
As the latent table is learned in parallel with the NeRF model parameters, we can treat this process as a fine-tuning optimization for an additional row in the latent table.
This row is initialized to the mean $\latentmean$ over the existing rows of the latent table, and is optimized using the same losses and optimizer as the main model.
Results for this fitting method are shown in \Section{results}.

\subsection{Unconditional Generation}
\label{sec:unconditional}
To sample novel objects from the space learned by our model, we sample latent codes from the empirical distribution $\latentdistribution$ defined by the rows of the latent table $\latenttable$.
We model $\latentdistribution$ as a multivariate Gaussian with mean $\latentmean$ and covariance $\latentcovariance$ found by performing principal component analysis on the rows of $\latenttable$.
Similar to other generative models which use a Gaussian prior on their latent variables, we observe a trade-off between diversity and quality of samples when sampling further away from the mean of the distribution.
Thus, we employ the ``truncation trick'' commonly used in the GAN literature\cite{Karras2019stylegan, marchesi2017megapixel, brock2018large, kingma2018glow} to control this trade-off.

\section{Results}
\label{sec:results}
\begin{figure*}[t]
\centering
\includegraphics[width=.95\linewidth]{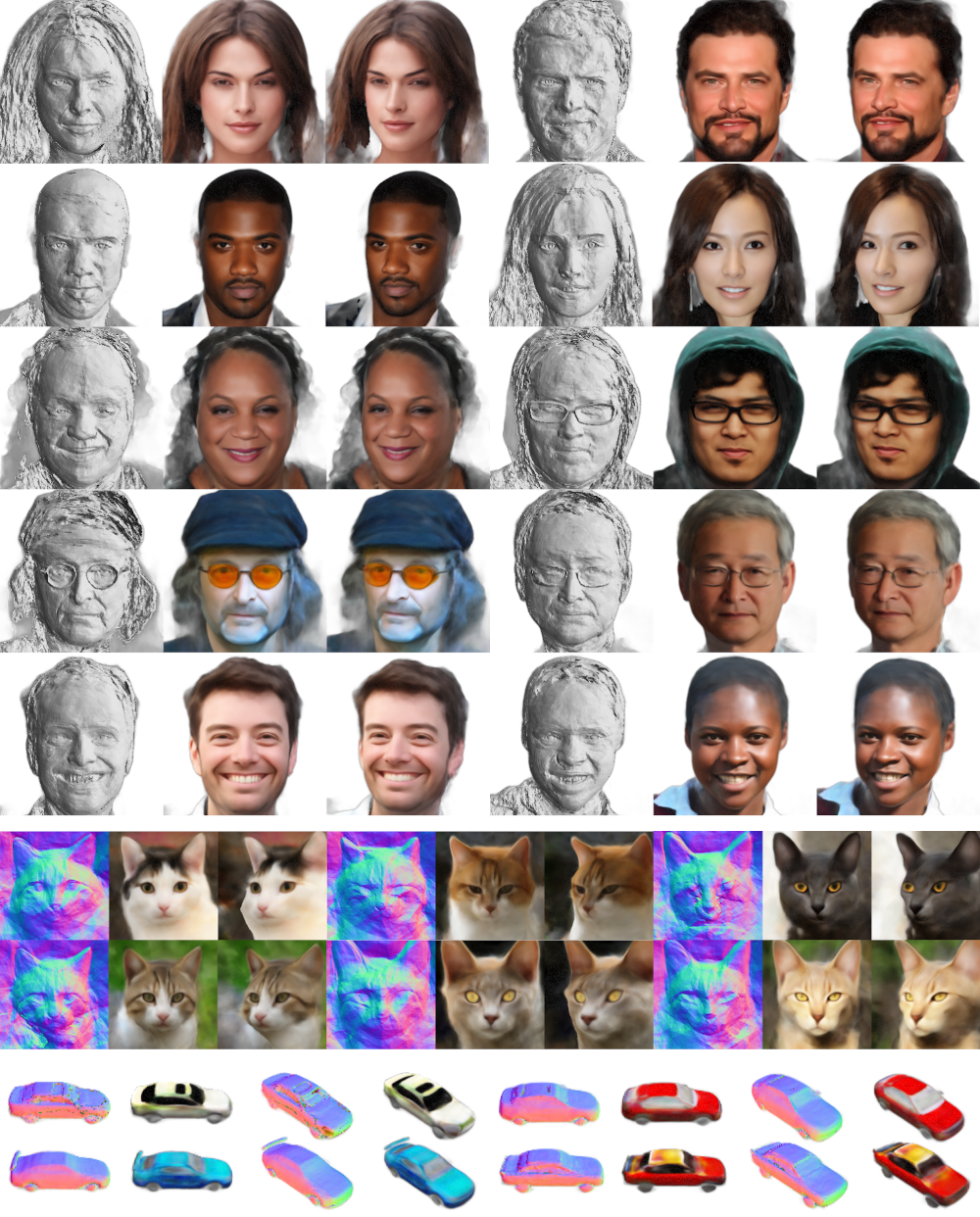}
\caption{
\textbf{Qualitative renders} --
Outputs from our method on all datasets.
Rows 1-2: CelebA-HQ~\cite{karras2017progressive},
Rows 3-5: FFHQ~\cite{Karras2019stylegan},
Rows 6-7: AFHQ Cats~\cite{choi2020stargan},
Rows 8-9: SRN Cars~\cite{Sitzmann19neurips_srn}.
}
\label{fig:qual_renders}
\end{figure*}

We visualize images rendered from our models trained on the 
CelebA-HQ~\cite{karras2017progressive}, FFHQ~\cite{Karras2019stylegan}, 
AFHQ~\cite{choi2020stargan}, 
and SRN Cars~\cite{Sitzmann19neurips_srn} datasets in \Figure{qual_renders}.
We include additional qualitative results from these datasets in the \SupplementaryMaterial.
To provide quantitative evaluation of our method and comparison to state of the art we perform a number of experiments, described in the following subsections.

\begin{table}
\small
\centering
\resizebox{\linewidth}{!}{ %
\begin{tabular}{@{}l|cccc@{}}
\toprule
Method & PSNR$\uparrow$ & SSIM$\uparrow$ & LPIPS$\downarrow$ & Res.\\
\midrule
\pigan (CelebA)        & 23.5 & 0.858 & 0.226 & \multirow{3}{*}{$128^2$}\\
Ours (FFHQ) & 29.0 & 0.913 & 0.199 \\
Ours (CelebA-HQ) & \bf{29.1} & \bf{0.914} & \bf{0.197} \\
\bottomrule
\end{tabular}
}
\caption{
\textbf{Reconstructions of training images} -- 
Metrics on a subset of 200 images from the \pigan~training set.
Our model achieves significantly higher reconstruction quality, regardless of whether it is trained on (FFHQ) or (CelebA-HQ).
}
\label{tab:quant_train_recon}
\end{table}
\begin{table}
\small
\centering
\resizebox{\linewidth}{!}{ %
\begin{tabular}{@{}l|ccccc@{}}
\toprule
Method & PSNR$\uparrow$ & SSIM$\uparrow$ & LPIPS$\downarrow$ & Res.\\
\midrule
\pigan (CelebA)        & 21.8 & 0.796 & 0.412 & \multirow{2}{*}{$256^2$}\\
Ours (CelebA-HQ) & \bf{26.2} & \bf{0.856} & \bf{0.363} \\
\midrule
\pigan (CelebA)        & 20.9 & 0.795 & 0.522 & \multirow{3}{*}{$512^2$}\\
Ours (CelebA-HQ) & 25.1 & 0.831 & 0.501 \\
Ours (FFHQ) & \bf{25.3} & \bf{0.836} & \bf{0.491} \\
\bottomrule
\end{tabular}
} %
\caption{
\textbf{Reconstructions of test images} -- 
Reconstruction quality (rows 1 and 2) of models trained on (CelebA) and (CelebA-HQ) on images from a 200-image subset of FFHQ, and (rows 3-5) of models trained at $256^2$ (Ours) and $128^2$ (\pigan) on high resolution $512^2$ versions of the test images.
}
\label{tab:quant_test_recon}
\end{table}
\subsection{Image Reconstruction -- \Table{quant_train_recon} and \Table{quant_test_recon}}
As \methodname is trained with an image reconstruction metric, we first perform experiments to evaluate how well images from the training dataset are reconstructed.
In \Table{quant_train_recon}, we show the average image reconstruction quality of both our method and \pigan for a 200-image subset of the \pigan training set (CelebA~\cite{liu2015deep}), as measured by peak signal to noise ratio (PSNR), structural similarity index measure (SSIM), and learned perceptual image patch similarity (LPIPS).~$\quad$~To compare against~\pigan, which does not learn latent codes corresponding to training images, we use the procedure included with the original~\pigan implementation for fitting images through test-time latent optimization.
Because this assumes perfectly forward facing pose, to make the comparison fair, we augment it with our camera fitting method to improve its results on profile-view images.
We also perform a more direct comparison of image fitting by testing on a set of held out images \textit{not seen} by the network during training.
To do this, we sample a set of 200 images from the FFHQ dataset and use the latent optimization procedure described in \Section{fitting} to produce reconstructions using a model trained on CelebA images.
We show the reconstruction metrics for these images using \methodname and \pigan in \Table{quant_test_recon} and examples of the reconstructed images in \Figure{supp_fitting}.

\begin{table}
  \small
  \centering
  \begin{tabular}{@{}l|ccc@{}}
    \toprule
    Method & PSNR$\uparrow$ & SSIM$\uparrow$ & Mask-LPIPS$\downarrow$\\
    \midrule
    \pigan (CelebA)                   & 24.5 & 0.918 & 0.102 \\
    Ours (CelebA-HQ) & 26.6 & 0.930 & 0.0989 \\
    Ours (FFHQ)      & \bf{27.0} & \bf{0.931} & \bf{0.0975} \\
    \bottomrule
  \end{tabular}
\caption{
\textbf{Novel view synthesis} --
We sample pairs of images from one frame for each subject in the HUMBI dataset~\cite{Yu2020humbi} and use them as query/target pairs.
The query image is used to optimize a latent representation of the subject's face, which is then rendered from the target view.
To evaluate how well the models have learned the 3D structure of faces, we then evaluate image reconstruction metrics for the face pixels of the predicted and target images after applying a mask computed from face landmarks.
}
\label{tab:quant_novel_view}
\end{table}
\subsection{Novel View Synthesis -- \Table{quant_novel_view}}
To evaluate the accuracy of the learned 3D structure, we perform image reconstruction experiments for synthesized novel views.
We render these novel views by performing image fitting on single frames from a synchronized multi-view face dataset, Human Multiview Behavioural Imaging~(HUMBI)~\cite{Yu2020humbi}, and reconstructing images using the camera parameters from other ground truth views of the same person.
The results of this experiment for \methodname and~\pigan are given in~\Table{quant_novel_view}.
We find that our model achieves significantly better reconstruction from novel views, indicating that our method has indeed learned a better 3D shape space than \pigan ~--~a shape space that is capable of generalizing to unseen data and does more than simply reproducing the query image from the query view.
We also show qualitative examples of novel views rendered by \methodname and \pigan in Figures \ref{fig:qual_novel_view} and \ref{fig:qual_fitting}.

\begin{table}
  \small
  \centering
  \begin{tabular}{@{}l|c@{}}
    \toprule
    Method & Depth Correlation$\uparrow$ \\
    \midrule
    Ground Truth & 66 \\
    \midrule
    AIGN (supervised) \cite{tung2017adversarial} & 50.81 \\
    DepthNetGAN (supervised) \cite{moniz2018unsupervised} & \bf{58.68} \\
    \midrule
    MOFA \cite{tewari2017mofa} & 15.97 \\
    DepthNet \cite{moniz2018unsupervised} & 35.77 \\
    \symmetry & 54.65 \\
    \midrule
    Ours (CelebA-HQ) & 50.18 \\
    \bottomrule
  \end{tabular}
\caption{
\textbf{Depth prediction} --
Correlation between predicted and true keypoint depth values on 3DFAW.
We compare to results from supervised and unsupervised methods as reported in \cite{Wu2020ulp} and \cite{moniz2018unsupervised}.
}
\label{tab:quant_depth_pred}
\end{table}
\subsection{Depth Prediction}

We also evaluate our shape model by predicting depth values for images where ground truth depth is available.
We use the 3DFAW dataset \cite{Yin2008bu4dfe, Zhang2013bp4d, Jeni2017dtf, Gross2010multipie, Jeni20163dfaw} which provides ground truth 3D keypoint locations.
For this task we fit latent codes from our model on the 3DFAW images and sample the predicted depth values for each image-space landmark location.
We use the same procedure as~\cite{Wu2020ulp} to compute the correlation of the predicted and ground truth depth values, which is recorded in in \Table{quant_depth_pred}.
While our score is not as high as the best performing unsupervised method (\cite{Wu2020ulp}), it outperforms several supervised and unsupervised methods specifically designed for depth prediction.

\subsection{High Resolution Image Synthesis}

To demonstrate the benefits of being able to train directly on high-resolution images, we quantitatively and qualitatively compare high-resolution renders from \methodname trained on 256$\times$256 FFHQ and CelebA-HQ images to those of \pigan trained on 128$\times$128 CelebA images (the largest feasible size used due to compute constraints).
These results are shown in \Table{quant_test_recon} and the \SupplementaryMaterial.
We find that for this task our models do much better at reproducing high-resolution detail, even though both methods are capable of producing ``infinite resolution'' images in theory.

\section{Discussion}

\subsection{Limitations and Future Work}
\begin{figure}[t]
\centering
\includegraphics[width=\linewidth]{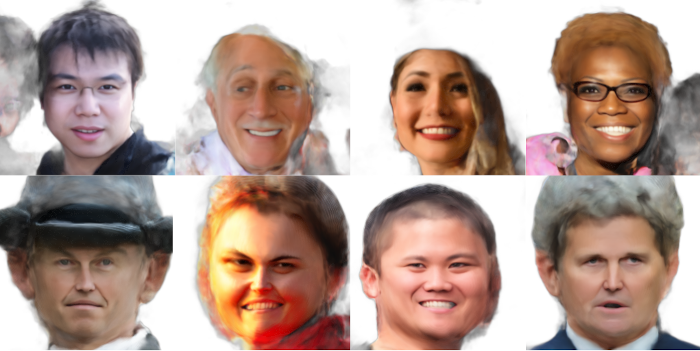}
\caption{
\textbf{Failure cases} --
For some images, the segmentation module fails to remove foreground objects that are not the subject, or erroneously removes parts of the subject (Top).
Our camera fitting also sometimes produces unreasonable 
camera estimates with focal lengths that are too short, or angles that are far from correct, 
resulting in distorted reconstruction of that subject during training
(Bottom).
Both failure modes occur in a small fraction our data, but the model still learns a space of 3D shape and appearance that performs well on structure-sensitive tasks, evident by our novel view synthesis (\Table{quant_novel_view}) and depth prediction (\Table{quant_depth_pred}) results.
}
\label{fig:qual_failure_cases}
\end{figure}

While our method produces very high-quality results from training on in-the-wild data, it is still reliant on other methods for extracting semantic information (i.e. landmarks) about the objects being observed.
As shown in \Figure{qual_failure_cases}, this dependence can lead to failure cases for objects where the estimated pose or segmentation is incorrect.
Finding a method to achieve alignment without any prior knowledge remains an open research question.

Also, while our auto-decoder framework has many advantages over GANs, it does not provide the same ability to maximize "plausibility" of the rendered images, which can result in some loss of detail.
A possible direction for future work would be to augment our method with adversarial training to further improve the perceptual quality of images rendered from novel latent codes.

\subsection{Ethical Considerations}
Our research on image generation focuses on socially beneficial use cases and applications. When developed correctly, generative models can do that in a number of ways including simulating a diverse population of users (fairness) and amplify the effectiveness of personal data thus reducing the need for large scale data collection (privacy). However, we acknowledge the potential for misuse and importance of acting responsibly. To that end, we will only release the code for reproducibility purposes, but will not  release any trained generative model.

\subsection{Conclusions}

We have presented \methodname, a method for learning spaces of 3D shape and appearance from datasets of single-view images.
Our experiments have shown that this method is able to learn effectively from unstructured, "in-the-wild" data, without incurring the high cost of a full-image discriminator, and while avoiding problems such as mode-dropping that are inherent to adversarial methods.

\section*{Acknowledgements}
We thank Matthew Brown, Kevin Swersky, David Fleet, and Viral Carpenter for their helpful technical insights and feedback.
Also Danica Matthews for her assistance with feline data collection.
This work was partly supported by the Natural Sciences and Engineering Research Council of Canada (NSERC) and by Compute Canada.

{
    \footnotesize
    \bibliographystyle{ieee_fullname}
    \bibliography{macros,main,nerf}

\begin{thebibliography}{10}\itemsep=-1pt

\bibitem{Andrew2001mvg}
Alex~M Andrew.
\newblock Multiple view geometry in computer vision.
\newblock {\em Kybernetes}, 2001.

\bibitem{Azuri2020gli}
Idan Azuri and Daphna Weinshall.
\newblock Generative latent implicit conditional optimization when learning
  from small sample.
\newblock In {\em ICPR}, 2020.

\bibitem{Barron21iccv_Mip_NeRF}
Jonathan Barron, Ben Mildenhall, Matthew Tancik, Peter Hedman, Ricardo
  Martin-Brualla, and Pratul Srinivasan.
\newblock {Mip-NeRF: A Multiscale Representation for Anti-Aliasing Neural
  Radiance Fields}.
\newblock In {\em ICCV}, October 2021.

\bibitem{binkowski2018demystifying}
Miko{\l}aj Bi{\'n}kowski, Danica~J Sutherland, Michael Arbel, and Arthur
  Gretton.
\newblock Demystifying mmd gans.
\newblock In {\em International Conference on Learning Representations}, 2018.

\bibitem{Blanz1999mms}
Volker Blanz and Thomas Vetter.
\newblock A morphable model for the synthesis of 3d faces.
\newblock In {\em Siggraph}, 1999.

\bibitem{Bojanowski2019ols}
Piotr Bojanowski, Armand Joulin, David Lopez-Pas, and Arthur Szlam.
\newblock Optimizing the {Latent} {Space} of {Generative} {Networks}.
\newblock In {\em Proceedings of the 35th {International} {Conference} on
  {Machine} {Learning}}, pages 600--609. PMLR, July 2018.
\newblock ISSN: 2640-3498.

\bibitem{borji2019pros}
Ali Borji.
\newblock Pros and cons of gan evaluation measures.
\newblock {\em Computer Vision and Image Understanding}, 179:41--65, 2019.

\bibitem{brock2018large}
Andrew Brock, Jeff Donahue, and Karen Simonyan.
\newblock Large scale {GAN} training for high fidelity natural image synthesis.
\newblock In {\em International Conference on Learning Representations}, 2019.

\bibitem{Chan21cvpr_piGAN}
Eric Chan, Marco Monteiro, Peter Kellnhofer, Jiajun Wu, and Gordon Wetzstein.
\newblock {pi-GAN: Periodic Implicit Generative Adversarial Networks for
  3D-Aware Image Synthesis}.
\newblock In {\em CVPR}, 2021.

\bibitem{Chen20cvpr_bsp}
Zhiqin Chen, Andrea Tagliasacchi, and Hao Zhang.
\newblock {BSP-Net: Generating Compact Meshes via Binary Space Partitioning}.
\newblock In {\em CVPR}, 2020.

\bibitem{Chen2019imnet}
Zhiqin Chen and Hao Zhang.
\newblock Learning implicit fields for generative shape modeling.
\newblock In {\em CVPR}, pages 5939--5948, 2019.

\bibitem{choi2020stargan}
Yunjey Choi, Youngjung Uh, Jaejun Yoo, and Jung-Woo Ha.
\newblock Stargan v2: Diverse image synthesis for multiple domains.
\newblock In {\em CVPR}, pages 8188--8197, 2020.

\bibitem{Dellaert2021nerfexplosion}
Frank Dellaert and Yen{-}Chen Lin.
\newblock Neural volume rendering: Nerf and beyond.
\newblock {\em CoRR}, abs/2101.05204, 2021.

\bibitem{Deng20cvpr_cvxnet}
Boyang Deng, Kyle Genova, Sofien Bouaziz, Geoffrey Hinton, Andrea Tagliasacchi,
  and Soroosh Yazdani.
\newblock {CvxNet: Learnable Convex Decomposition}.
\newblock In {\em CVPR}, 2020.

\bibitem{Egger2020tdm}
Bernhard Egger, William~AP Smith, Ayush Tewari, Stefanie Wuhrer, Michael
  Zollhoefer, Thabo Beeler, Florian Bernard, Timo Bolkart, Adam Kortylewski,
  Sami Romdhani, et~al.
\newblock 3d morphable face models—past, present, and future.
\newblock {\em ACM Transactions on Graphics (TOG)}, 39(5):1--38, 2020.

\bibitem{Gafni21cvpr_DNRF}
Guy Gafni, Justus Thies, Michael Zollhöfer, and Matthias Nießner.
\newblock {Dynamic Neural Radiance Fields for Monocular 4D Facial Avatar
  Reconstruction}.
\newblock In {\em CVPR}, 2021.

\bibitem{Gao21iccv_DynamicVS}
Chen Gao, Ayush Saraf, Johannes Kopf, and Jia-Bin Huang.
\newblock {Dynamic View Synthesis from Dynamic Monocular Video}.
\newblock In {\em ICCV}, October 2021.

\bibitem{Genova19iccv_sif}
Kyle Genova, Forrester Cole, Daniel Vlasic, Aaron Sarna, William Freeman, and
  Thomas Funkhouser.
\newblock {Learning Shape Templates with Structured Implicit Functions}.
\newblock In {\em ICCV}, 2019.

\bibitem{Gerig2018mfm}
Thomas Gerig, Andreas Morel-Forster, Clemens Blumer, Bernhard Egger, Marcel
  Luthi, Sandro Sch{\"o}nborn, and Thomas Vetter.
\newblock Morphable face models-an open framework.
\newblock In {\em 2018 13th IEEE International Conference on Automatic Face \&
  Gesture Recognition (FG 2018)}, pages 75--82. IEEE, 2018.

\bibitem{Gotardo2018pdf}
Paulo Gotardo, Jérémy Riviere, Derek Bradley, Abhijeet Ghosh, and Thabo
  Beeler.
\newblock Practical dynamic facial appearance modeling and acquisition.
\newblock In {\em Siggraph Asia}, 2018.

\bibitem{Gross2010multipie}
Ralph Gross, Iain Matthews, Jeffrey Cohn, Takeo Kanade, and Simon Baker.
\newblock Multi-pie.
\newblock {\em Image and vision computing}, 28(5):807--813, 2010.

\bibitem{Anonymous2021stylenerf}
Jiatao Gu, Lingjie Liu, Peng Wang, and Christian Theobalt.
\newblock Stylenerf: A style-based 3d aware generator for high-resolution image
  synthesis.
\newblock In {\em International Conference on Learning Representations}, 2022.

\bibitem{Henzler21cvpr_object_categories}
Philipp Henzler.
\newblock {Unsupervised Learning of 3D Object Categories from Videos in the
  Wild}.
\newblock In {\em CVPR}, 2021.

\bibitem{heusel2017gans}
Martin Heusel, Hubert Ramsauer, Thomas Unterthiner, Bernhard Nessler, and Sepp
  Hochreiter.
\newblock Gans trained by a two time-scale update rule converge to a local nash
  equilibrium.
\newblock {\em Advances in neural information processing systems}, 30, 2017.

\bibitem{Jang21iccv_CodeNeRF}
Wongbong Jang and Lourdes Agapito.
\newblock {CodeNeRF: Disentangled Neural Radiance Fields for Object
  Categories}.
\newblock In {\em ICCV}, October 2021.

\bibitem{Jeni2017dtf}
László~A. Jeni, Jeffrey~F. Cohn, and Takeo Kanade.
\newblock Dense {3D} face alignment from {2D} video for real-time use.
\newblock {\em Image and Vision Computing}, 58:13--24, 2016.
\newblock doi:10.1016/j.imavis.2016.05.009.

\bibitem{Jeni20163dfaw}
László~A. Jeni, Sergey Tulyakov, Lijun Yin, Nicu Sebe, and Jeffrey~F. Cohn.
\newblock {The First 3D Face Alignment in the Wild (3DFAW) Challenge}.
\newblock In {\em {ECCVW}}, 2016.

\bibitem{Kajiya1984rtv}
James~T. Kajiya and Brian~P. Von~Herzen.
\newblock Ray tracing volume densities.
\newblock In {\em Siggraph}, 1984.

\bibitem{karras2017progressive}
Tero Karras, Timo Aila, Samuli Laine, and Jaakko Lehtinen.
\newblock Progressive {Growing} of {GANs} for {Improved} {Quality},
  {Stability}, and {Variation}.
\newblock In {\em ICLR}, 2018.

\bibitem{Karras2019stylegan}
Tero Karras, Samuli Laine, and Timo Aila.
\newblock A style-based generator architecture for generative adversarial
  networks.
\newblock In {\em CVPR}, 2019.

\bibitem{kingma2015adam}
Diederik~P Kingma and Jimmy Ba.
\newblock Adam: A method for stochastic optimization.
\newblock In {\em ICLR (Poster)}, 2015.

\bibitem{kingma2018glow}
Durk~P Kingma and Prafulla Dhariwal.
\newblock Glow: {Generative} {Flow} with {Invertible} 1x1 {Convolutions}.
\newblock In {\em NeurIPS}, 2018.

\bibitem{levenberg1944method}
Kenneth Levenberg.
\newblock A method for the solution of certain non-linear problems in least
  squares.
\newblock {\em Quarterly of applied mathematics}, 2(2):164--168, 1944.

\bibitem{Yin2008bu4dfe}
{Lijun Yin}, {Xiaochen Chen}, {Yi Sun}, {Tony Worm}, and {Michael Reale}.
\newblock A {High}-{Resolution} {3D} {Dynamic} {Facial} {Expression}
  {Database}.
\newblock In {\em The 8th {International} {Conference} on {Automatic} {Face}
  and {Gesture} {Recognition} ({FGR08})}, Amsterdam, The Netherlands, 2008.

\bibitem{Lin21iccv_BARF}
Chen-Hsuan Lin, Wei-Chiu Ma, Antonio Torralba, and Simon Lucey.
\newblock {BARF: Bundle-Adjusting Neural Radiance Fields}.
\newblock In {\em ICCV}, October 2021.

\bibitem{liu2015deep}
Ziwei Liu, Ping Luo, Xiaogang Wang, and Xiaoou Tang.
\newblock Deep learning face attributes in the wild.
\newblock In {\em CVPR}, 2015.

\bibitem{Malik2016threer}
Jitendra Malik, Pablo Arbel{\'a}ez, Jo{\~a}o Carreira, Katerina Fragkiadaki,
  Ross Girshick, Georgia Gkioxari, Saurabh Gupta, Bharath Hariharan, Abhishek
  Kar, and Shubham Tulsiani.
\newblock The three {R}’s of computer vision: Recognition, reconstruction and
  reorganization.
\newblock {\em Pattern Recognition Letters}, 72:4--14, 2016.

\bibitem{marchesi2017megapixel}
Marco Marchesi.
\newblock Megapixel size image creation using generative adversarial networks.
\newblock {\em arXiv preprint arXiv:1706.00082}, 2017.

\bibitem{MartinBrualla21cvpr_nerfw}
Ricardo Martin-Brualla, Noha Radwan, Mehdi S.~M. Sajjadi, Jonathan Barron,
  Alexey Dosovitskiy, and Daniel Duckworth.
\newblock {NeRF in the Wild: Neural Radiance Fields for Unconstrained Photo
  Collections}.
\newblock In {\em CVPR}, 2021.

\bibitem{Google2021fm}
{MediaPipe Face Mesh}.
\newblock \url{https://google.github.io/mediapipe/solutions/face_mesh.html}.
\newblock Accessed: 2021-10-20.

\bibitem{Google2021ss}
{MediaPipe Selfie Segmentation}.
\newblock
  \url{https://google.github.io/mediapipe/solutions/selfie_segmentation.html}.
\newblock Accessed: 2021-10-20.

\bibitem{Meng21iccv_GNeRF}
Quan Meng.
\newblock {GNeRF: GAN-Based Neural Radiance Field Without Posed Camera}.
\newblock In {\em ICCV}, October 2021.

\bibitem{Mescheder19cvpr_occupancy_net}
Lars Mescheder, Michael Oechsle, Michael Niemeyer, Sebastian Nowozin, and
  Andreas Geiger.
\newblock {Occupancy Networks: Learning 3D Reconstruction in Function Space}.
\newblock In {\em CVPR}, 2019.

\bibitem{Mildenhall20eccv_nerf}
Ben Mildenhall, Pratul Srinivasan, Matthew Tancik, Jonathan Barron, Ravi
  Ramamoorthi, and Ren Ng.
\newblock {NeRF: Representing Scenes as Neural Radiance Fields for View
  Synthesis}.
\newblock In {\em ECCV}, pages 405--421. Springer, 2020.

\bibitem{moniz2018unsupervised}
Joel Ruben~Antony Moniz, Christopher Beckham, Simon Rajotte, Sina Honari, and
  Chris Pal.
\newblock Unsupervised {Depth} {Estimation}, {3D} {Face} {Rotation} and
  {Replacement}.
\newblock In {\em NeurIPS}, 2018.

\bibitem{nguyen2019hologan}
Thu Nguyen-Phuoc, Chuan Li, Lucas Theis, Christian Richardt, and Yong-Liang
  Yang.
\newblock Hologan: Unsupervised learning of 3d representations from natural
  images.
\newblock In {\em Proceedings of the IEEE/CVF International Conference on
  Computer Vision}, pages 7588--7597, 2019.

\bibitem{Niemeyer2021campari}
Michael Niemeyer and Andreas Geiger.
\newblock {CAMPARI}: Camera-aware decomposed generative neural radiance fields.
\newblock In {\em International Conference on 3D Vision (3DV)}, 2021.

\bibitem{Niemeyer21cvpr_GIRAFFE}
Michael Niemeyer and Andreas Geiger.
\newblock {GIRAFFE: Representing Scenes as Compositional Generative Neural
  Feature Fields}.
\newblock In {\em CVPR}, 2021.

\bibitem{Oechsle21iccv_UNISURF}
Michael Oechsle, Songyou Peng, and Andreas Geiger.
\newblock {UNISURF: Unifying Neural Implicit Surfaces and Radiance Fields for
  Multi-View Reconstruction}.
\newblock In {\em ICCV}, October 2021.

\bibitem{Pan2021shadegan}
Xingang Pan, Xudong Xu, Chen~Change Loy, Christian Theobalt, and Bo Dai.
\newblock A shading-guided generative implicit model for shape-accurate
  3d-aware image synthesis.
\newblock In {\em Advances in Neural Information Processing Systems (NeurIPS)},
  2021.

\bibitem{Park19cvpr_deepsdf}
Jeong~Joon Park, Pete Florence, Julian Straub, Richard Newcombe, and Steven
  Lovegrove.
\newblock {DeepSDF: Learning Continuous Signed Distance Functions for Shape
  Representation}.
\newblock In {\em CVPR}, 2019.

\bibitem{Park21iccv_nerfies}
Keunhong Park, Utkarsh Sinha, Jonathan Barron, Sofien Bouaziz, Dan Goldman,
  Steven Seitz, and Ricardo Martin-Brualla.
\newblock {Nerfies: Deformable Neural Radiance Fields}.
\newblock In {\em ICCV}, October 2021.

\bibitem{Park2021hypernerf}
Keunhong Park, Utkarsh Sinha, Peter Hedman, Jonathan~T. Barron, Sofien Bouaziz,
  Dan~B Goldman, Ricardo Martin-Brualla, and Steven~M. Seitz.
\newblock Hypernerf: A higher-dimensional representation for topologically
  varying neural radiance fields.
\newblock {\em ACM Trans. Graph.}, 2021.

\bibitem{Paysan2009tdf}
Pascal Paysan, Reinhard Knothe, Brian Amberg, Sami Romdhani, and Thomas Vetter.
\newblock A 3d face model for pose and illumination invariant face recognition.
\newblock In {\em {IEEE International Conference on Advanced Video and Signal
  Based Surveillance}}, 2009.

\bibitem{Riviere2020ssh}
Jérémy Riviere, Paulo Gotardo, Derek Bradley, Abhijeet Ghosh, and Thabo
  Beeler.
\newblock Single-shot high-quality facial geometry and skin appearance capture.
\newblock In {\em Siggraph}, 2020.

\bibitem{salimans2016improved}
Tim Salimans, Ian Goodfellow, Wojciech Zaremba, Vicki Cheung, Alec Radford, and
  Xi Chen.
\newblock Improved techniques for training gans.
\newblock {\em Advances in neural information processing systems},
  29:2234--2242, 2016.

\bibitem{Schwarz20neurips_graf}
Katja Schwarz, Yiyi Liao, Michael Niemeyer, and Andreas Geiger.
\newblock {GRAF: Generative Radiance Fields for 3D-Aware Image Synthesis}.
\newblock In {\em Adv. Neural Inform. Process. Syst.}, 2020.

\bibitem{NEURIPS2020_e92e1b47}
Katja Schwarz, Yiyi Liao, Michael Niemeyer, and Andreas Geiger.
\newblock Graf: Generative radiance fields for 3d-aware image synthesis.
\newblock In H. Larochelle, M. Ranzato, R. Hadsell, M.~F. Balcan, and H. Lin,
  editors, {\em Advances in Neural Information Processing Systems}, volume~33,
  pages 20154--20166. Curran Associates, Inc., 2020.

\bibitem{Singh2019gloss}
Sidak~Pal Singh, Angela Fan, and Michael Auli.
\newblock {GLOSS}: Generative latent optimization of sentence representations.
\newblock {\em arXiv preprint arXiv:1907.06385}, 2019.

\bibitem{Sitzmann20neurips_metaSDF}
Vincent Sitzmann, Eric~R. Chan, Richard Tucker, Noah Snavely, and Gordon
  Wetzstein.
\newblock {MetaSDF}: Meta-learning signed distance functions.
\newblock In {\em NeurIPS}, 2020.

\bibitem{Sitzmann2021lfn}
Vincent Sitzmann, Semon Rezchikov, William~T. Freeman, Joshua~B. Tenenbaum, and
  Fredo Durand.
\newblock Light field networks: Neural scene representations with
  single-evaluation rendering.
\newblock In {\em Proc. NeurIPS}, 2021.

\bibitem{Sitzmann19neurips_srn}
Vincent Sitzmann, Michael Zollhöfer, and Gordon Wetzstein.
\newblock {Scene Representation Networks: Continuous 3D-Structure-Aware Neural
  Scene Representations}.
\newblock In {\em Adv. Neural Inform. Process. Syst.}, 2019.

\bibitem{suwajanakorn2018discovery}
Supasorn Suwajanakorn, Noah Snavely, Jonathan~J Tompson, and Mohammad Norouzi.
\newblock Discovery of latent 3d keypoints via end-to-end geometric reasoning.
\newblock {\em Advances in Neural Information Processing Systems}, 31, 2018.

\bibitem{Tan1995rdd}
Shufeng Tan and Michael~L. Mayrovouniotis.
\newblock Reducing data dimensionality through optimizing neural network
  inputs.
\newblock {\em {AIChE} Journal}, 41(6):1471--1480, 1995.

\bibitem{tewari2017mofa}
Ayush Tewari, Michael Zollhofer, Hyeongwoo Kim, Pablo Garrido, Florian Bernard,
  Patrick Perez, and Christian Theobalt.
\newblock Mofa: Model-based deep convolutional face autoencoder for
  unsupervised monocular reconstruction.
\newblock In {\em Proceedings of the IEEE International Conference on Computer
  Vision Workshops}, pages 1274--1283, 2017.

\bibitem{Threedmd2021products}
{Products - 3dMD}.
\newblock \url{https://3dmd.com/products/}.
\newblock Accessed: 2021-10-25.

\bibitem{Trevithick21iccv_GRF}
Alex Trevithick and Bo Yang.
\newblock {GRF: Learning a General Radiance Field for 3D Scene Representation
  and Rendering}.
\newblock In {\em ICCV}, October 2021.

\bibitem{tung2017adversarial}
Hsiao-Yu~Fish Tung, Adam~W Harley, William Seto, and Katerina Fragkiadaki.
\newblock Adversarial inverse graphics networks: Learning 2d-to-3d lifting and
  image-to-image translation from unpaired supervision.
\newblock In {\em 2017 IEEE International Conference on Computer Vision
  (ICCV)}, pages 4364--4372. IEEE, 2017.

\bibitem{Wang21arxiv_NeRFminusminus}
Zirui Wang, Shangzhe Wu, Weidi Xie, Min Chen, and Victor~Adrian Prisacariu.
\newblock {NeRF--: Neural Radiance Fields Without Known Camera Parameters}.
\newblock {\em https://arxiv.org/abs/2102.07064}, 2021.

\bibitem{Wu2020ulp}
Shangzhe Wu, Christian Rupprecht, and Andrea Vedaldi.
\newblock Unsupervised learning of probably symmetric deformable 3d objects
  from images in the wild.
\newblock In {\em CVPR}, 2020.

\bibitem{Xie2021fignerf}
Christopher Xie, Keunhong Park, Ricardo Martin-Brualla, and Matthew Brown.
\newblock {FiG}-{NeRF}: Figure-ground neural radiance fields for 3d object
  category modelling.
\newblock {\em arXiv preprint arXiv:2104.08418}, 2021.

\bibitem{Yu21cvpr_pixelNeRF}
Alex Yu, Vickie Ye, Matthew Tancik, and Angjoo Kanazawa.
\newblock {pixelNeRF: Neural Radiance Fields from One or Few Images}.
\newblock In {\em CVPR}, 2021.

\bibitem{Yu2020humbi}
Zhixuan Yu, Jae~Shin Yoon, In~Kyu Lee, Prashanth Venkatesh, Jaesik Park, Jihun
  Yu, and Hyun~Soo Park.
\newblock {HUMBI}: A large multiview dataset of human body expressions.
\newblock In {\em CVPR}, 2020.

\bibitem{Zhang20arxiv_NeRFplusplus}
Kai Zhang, Gernot Riegler, Noah Snavely, and Vladlen Koltun.
\newblock {NERF++}: Analyzing and improving neural radiance fields.
\newblock {\em https://arxiv.org/abs/2010.07492}, 2020.

\bibitem{Zhang2013bp4d}
Xing Zhang, Lijun Yin, Jeffrey~F. Cohn, Shaun Canavan, Michael Reale, Andy
  Horowitz, and Peng Liu.
\newblock A high-resolution spontaneous {3D} dynamic facial expression
  database.
\newblock {\em Image and Vision Computing (special issue of The Best of Face
  and Gesture 2013)}, 32:692--706, 2014.

\end{thebibliography}
}

\clearpage

\appendix

\setcounter{page}{1}

\twocolumn[
\centering
\Large
\textbf{LOLNeRF: Learn from One Look} \\
\vspace{0.5em}Supplementary Material \\
\vspace{1.0em}
] %

\begin{figure}[b]
\centering
\begin{overpic}
[width=\linewidth]
{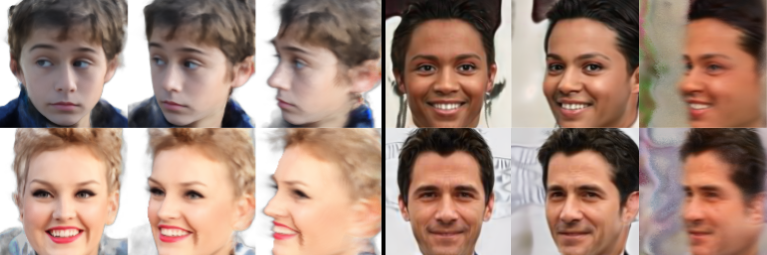}
\put(8,34){\footnotesize{$0^\circ$}}
\put(22,34){\footnotesize{$30^\circ$}}
\put(40,34){\footnotesize{$60^\circ$}}
\put(58,34){\footnotesize{$0^\circ$}}
\put(72,34){\footnotesize{$30^\circ$}}
\put(90,34){\footnotesize{$60^\circ$}}
\put(20,-3){\footnotesize{LOLNeRF}}
\put(70,-3){\footnotesize{$\pi$-GAN}}
\end{overpic}
\caption{
\textbf{High view angles} -- 
both methods show degraded quality at high angles, but ours maintains better sharpness and view-consistency.
}
\label{fig:supp_high_angle}
\end{figure}

\begin{figure*}[t]
\centering
\begin{overpic}
[width=\linewidth]
{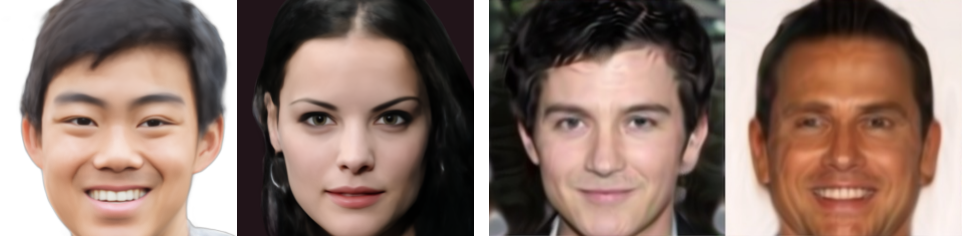}
\put(71,-2){\footnotesize\rotatebox{0}{\pigan}}
\put(20,-2){\footnotesize\rotatebox{0}{Our Method}}
\end{overpic}
\vspace{-1em}
\caption{
\textbf{High resolution renders} --
Our method is trained natively to reconstruct high resolution images and can reconstruct sharp details, while \pigan reconstructions lack detail beyond its training resolution and contain artifacts that become easily visible.
}
\label{fig:qual_highres}
\end{figure*}

\section{Qualitative Results -- Figures \ref{fig:supp_high_angle}, \ref{fig:qual_highres}}

We provide additional qualitative comparisons with \pigan in Figures \ref{fig:supp_high_angle} and \ref{fig:qual_highres}, which show novel views at high angles and high resolution renders, respectively.

Please also see the \dr{project website} (\url{https://lolnerf.github.io}) for animations which demonstrate the high-quality 3D structure and novel views that our method produces.

\section{Camera Fitting Procedure}

For a class-specific landmarker which provides estimates for $M$ 2D landmarks $\mathbf{\ell} \in \real^{M\times 2}$, we estimate the extrinsics $\mathbf{T}$ and (optionally) intrinsics $\mathbf{K}$ of a camera which minimizes the reprojection error between $\mathbf{\ell}$ and a set of canonical 3D positions $\mathbf{p} \in \real^{M\times 3}$.
We achieve this by solving the following least-squares optimization:
\begin{equation}
    \argmin_{\mathbf{T}, \mathbf{K}} ||\mathbf{\ell} - P(\mathbf{p} | \mathbf{T}, \mathbf{K})||^2
\end{equation}
where $P(\mathbf{x} | \mathbf{T}, \mathbf{K})$ represents the projection operation for a world-space position vector $\mathbf{x}$ into image space.
We perform this optimization using the Levenberg–Marquardt algorithm~\cite{levenberg1944method}.
The canonical positions $\mathbf{p}$ may be either manually specified or derived from data.
For human faces we use a predetermined set of positions which correspond to the known average geometry of the human face.
For AFHQ, we perform a version of the above optimization jointly across all images where $\mathbf{p}$ is also a free variable, and constrained only to obey symmetry.

In our experiments we predict camera intrinsics for human face data, but use fixed intrinsics for AFHQ where the landmarks are less effective in constraining the focal length. For SRN cars, we use the camera intrinsics and extrinsics provided with the dataset, though we note that semantically consistent landmarkers do exist for this class of data~\cite{suwajanakorn2018discovery}.

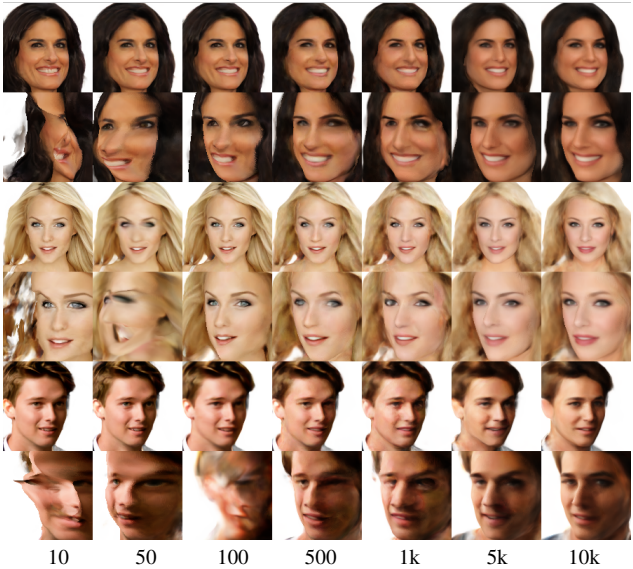
\begin{figure}
\centering
\begin{overpic}
[width=\linewidth]
{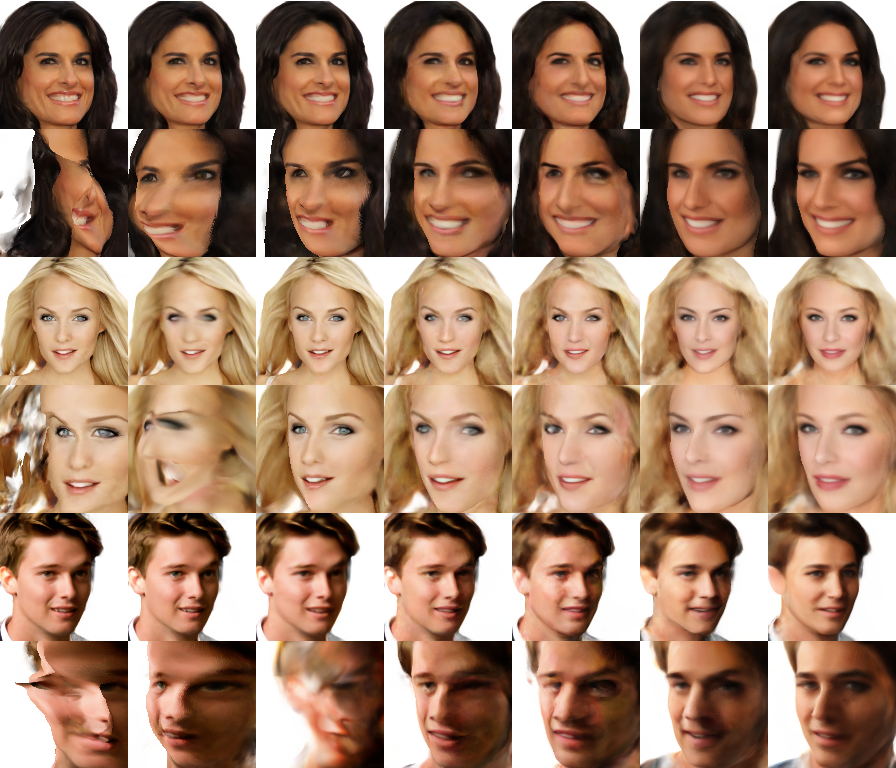}
\put(7,-4){\footnotesize 10}
\put(21,-4){\footnotesize 50}
\put(34,-4){\footnotesize 100}
\put(48,-4){\footnotesize 500}
\put(63,-4){\footnotesize 1k}
\put(77,-4){\footnotesize 5k}
\put(90,-4){\footnotesize 10k}
\end{overpic}
\vspace{0.1em}
\caption{
\textbf{Dataset size ablation} -- we show the behaviour of the method as the size of the training dataset varies.
Rows 1, 3, and 5 show the learned reconstruction of training images from the predicted view for that image.
Rows 2, 4, and 6 show the models rendered from a novel view.
The columns show the results as the total number of training images is increased from ten to ten thousand.
Note that this ablation is performed with lower-capacity networks (256 neurons per layer), both to avoid wasting energy in training and to more clearly show how reconstruction quality changes with dataset size.
}
\label{fig:data_ablation}
\end{figure}

\section{Dataset Size Ablation -- \Figure{data_ablation}}

To quantify our method's dependence on large amounts of data, we perform an ablation study in which we train models with subsets of the full dataset.
We then show the novel view synthesis quality from these networks as a way of determining how well they have generalized to reconstructing different views.
The results of this experiment are shown in~\Figure{data_ablation}.
We find a clear trade-off in quality of the training image reconstruction and quality of the learned 3D structure as the dataset size increases.
Very small datasets reconstruct their training images with high accuracy, but produce completely unreasonable geometry and novel views.
As the number of training images increases, the accuracy of reconstruction slowly decreases, but the predicted structure generalizes to become much more consistent and geometrically reasonable.

\begin{figure}[t]
\centering
\begin{overpic}[width=0.495\linewidth, trim=0 0 692 0, clip]
{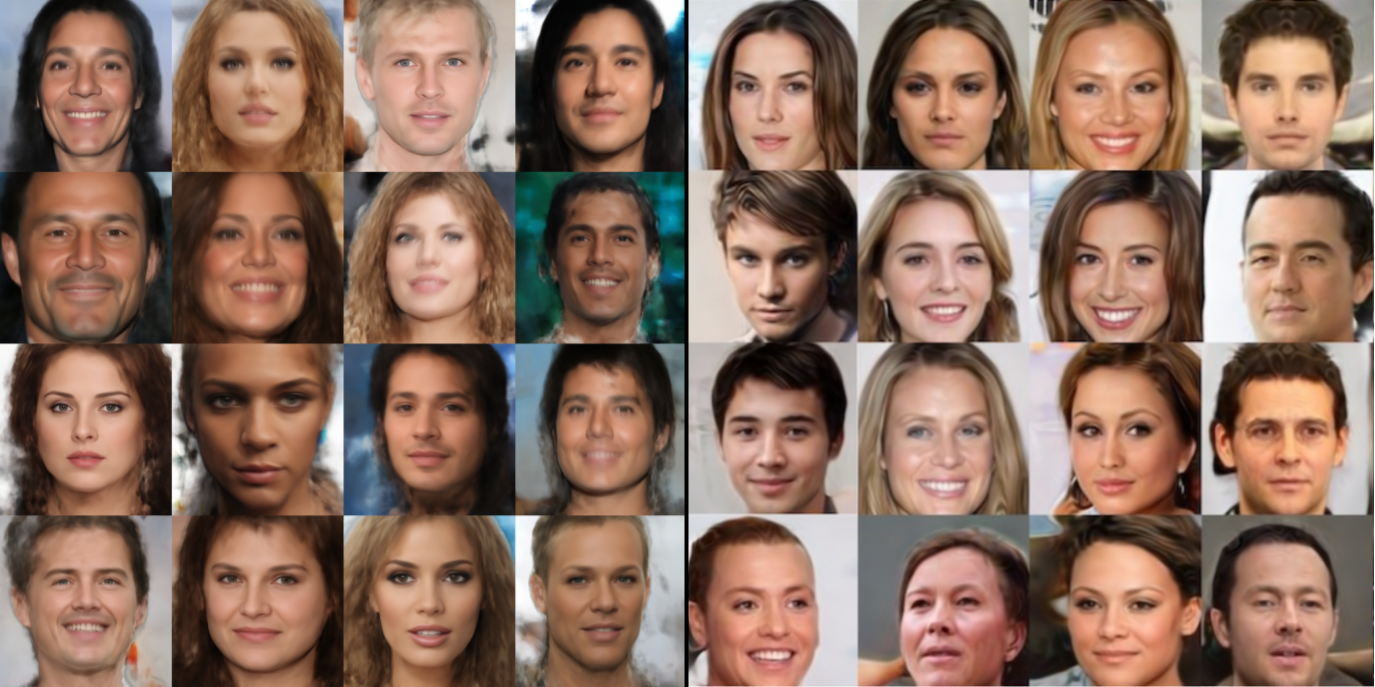}
\put(34,-8){\footnotesize{Our Method}}
\end{overpic}
\begin{overpic}[width=0.495\linewidth, trim=692 0 0 0, clip]
{fig/unconditional.png}
\put(34,-8){\footnotesize{\pigan}}
\end{overpic}
\vspace{-.5em}
\caption{
\textbf{Unconditional generation} --
Both methods produce samples that resemble the training distribution in appearance and shape.
Ours produce sharper details in the central face region where the data is more consistent, while \pigan produces more plausible hair and backgrounds due to its adversarial training.
}
\vspace{-.5em}
\label{fig:qual_unconditional}
\end{figure}

\section{Unconditional Sampling -- \Figure{qual_unconditional}}
\begin{table}
  \small
  \centering
  \begin{tabular}{@{}l|ccc@{}}
    \toprule
    Method & FID$\downarrow$ & KID$\downarrow$ & IS$\uparrow$\\
    \midrule
    HoloGAN~\cite{nguyen2019hologan} & 39.7 & 2.91 & 1.89\\
    GRAF~\cite{NEURIPS2020_e92e1b47} & 41.1 & 2.29 & 2.34\\
    \pigan & \bf{14.7} & 0.39 & \bf{2.62}\\
    Ours & 128.2 & \bf{0.11} & 2.34 \\
    \bottomrule
  \end{tabular}
  \caption{\textbf{Unconditional Sampling Quality} -- 
  Perceptual image distribution quality metrics on CelebA for our model and baselines.
  The results for HoloGAN and GRAF are taken from~\cite{Chan21cvpr_piGAN}.
  }
  \label{tab:quant_unconditional}
\end{table}

To evaluate the quality of unconditional samples generated by our PCA-based sampling method, we compute three standard quality metrics for generative image models on these renders: Frechet Inception Distace (FID)~\cite{heusel2017gans}, Kernel Inception Distance, (KID)~\cite{binkowski2018demystifying}, and Inception Score (IS)~\cite{salimans2016improved}, the results of which are shown in \Table{quant_unconditional}.
We find that our method achieves an inception score competitive with other 3D-aware GAN methods, indicating that we are able to model a variety of facial appearances.
Our result for the distribution distance metrics, FID and KID, however, show opposing results with our method doing far worse in FID but better in KID.
The reason for this is not entirely clear, but FID has been shown to be sensitive to noise~\cite{borji2019pros}, and details in the peripheral areas of our generated images show more noise-like artifacts than \pigan.
Regardless, we do not necessarily expect to outperform \pigan for these metrics as it is trained to produce images with high perceptual quality as determined by a CNN, which is a better proxy for these distribution metrics than our reconstruction loss.

\section{Architecture Details}

Our architecture uses a standard NeRF backbone architecture as described in~\cite{Mildenhall20eccv_nerf} with a few modifications.
In addition to the standard positional encoding we condition the network on an additional latent code by concatenating it alongside the positional encoding.
For SRN cars and AFHQ we use the standard 256 neuron network width and 256-dimensional latents for this network, but we increase to 1024 neurons and 2048-dimensional latents for our high-resolution CelebA-HQ and FFHQ models.
For our background model we use a 5-layer, 256-neuron relu MLP in all cases.
During training, we use 128 samples per ray for volume rendering with no hierarchical sampling.

\balance

\section{Training Details}

We train each model for 500k iterations using a batch size of 32 pixels per image, with a total of 4096 images included in each batch.
For comparison, with 256$^2$ images, this compute budget would allow just 2 images per batch for a GAN-based method which renders entire frames.

We train with an ADAM~\cite{kingma2015adam} optimizer using exponential decay for the learning rate from $5\times10^{-4}$ to $1\times10^{-4}$.
We run each training job using 64 v4 Tensor Processing Unit chips, taking approximately 36 hours to complete for our high resolution models.

\end{document}